\documentclass[letterpaper, 10 pt, conference]{ieeeconf}  

\IEEEoverridecommandlockouts                              

\let\labelindent\relax
\usepackage{amsmath} 
\usepackage{amssymb}  
\usepackage{cite}
\usepackage[caption=false]{subfig}     
\usepackage{enumitem}  
\usepackage{float}
\usepackage{afterpage}
\usepackage{stfloats}
\usepackage{tabularx}
\usepackage{array}
\usepackage{makecell}
\usepackage{graphicx}
\usepackage{booktabs}
\usepackage{multirow}
\usepackage{dsfont}
\usepackage{romannum}
\usepackage{hyperref}
\usepackage{xcolor}

\newcommand{\secref}[1]{Section~\ref{#1}}
\renewcommand{\eqref}[1]{(\ref{#1})}
\newcommand{\figref}[1]{Figure~\ref{#1}}
\newcommand{\subfig}[1]{\textit{#1}}
\newcommand{\tabref}[1]{Table~\ref{#1}}
\newcommand{\taskref}[1]{Task~\ref{#1}}

\newcommand{\ie}{\textit{i.e.}}
\newcommand{\eg}{\textit{e.g.}}

\def\inet{CNN}
\def\mfinet{MF-CNN}
\def\cnnlstm{CNN-LSTM}
\def\proposed{DECISION}
\def\Backbone{DECISION w/o L\#1-3}
\def\EarlyLayer{DECISION w/o L\#2-3}
\def\LateLayer{DECISION w/o L\#1-2}
\def\nobranch{DECISION w/o multimodal mem.}
\def\human{Human Teleoperation}
\def\fullname{\textbf{D}eep r\textbf{E}current \textbf{C}ontroller for v\textbf{IS}ual nav\textbf{I}gati\textbf{ON}}

\def\SR{SR}
\def\CR{CR}

\def\intentleft{\texttt{turn-left}}
\def\intentright{\texttt{turn-right}}
\def\intentforward{\texttt{go-forward}}
\def\intentelevator{\texttt{take-elevator}}

\def\branch{multimodal memory}

\newcommand{\cell}[1]{\#{#1}}
\def\video{\textcolor{magenta}{\url{https://adacomp.comp.nus.edu.sg/inet/}}}

\title{\LARGE \bf
Deep Visual Navigation under Partial Observability  
}

\newcommand\blfootnote[1]{%
\begingroup 
\renewcommand\thefootnote{}\footnote{#1}%
\addtocounter{footnote}{-1}%
\endgroup 
}

\author{
Bo Ai, 
Wei Gao, 
Vinay, and 
David Hsu, \IEEEmembership{Fellow, IEEE}
}

\begin{document}

\twocolumn[{%
\renewcommand\twocolumn[1][]{#1}%

\maketitle

\begin{figure}[H]
\hsize=\textwidth 
  \centering
  \setlength{\tabcolsep}{2pt} 
  \renewcommand{\arraystretch}{0.5} 
  \vspace{-20pt}
  \begin{tabular}{ccccc}
    \includegraphics[width=0.4\columnwidth]{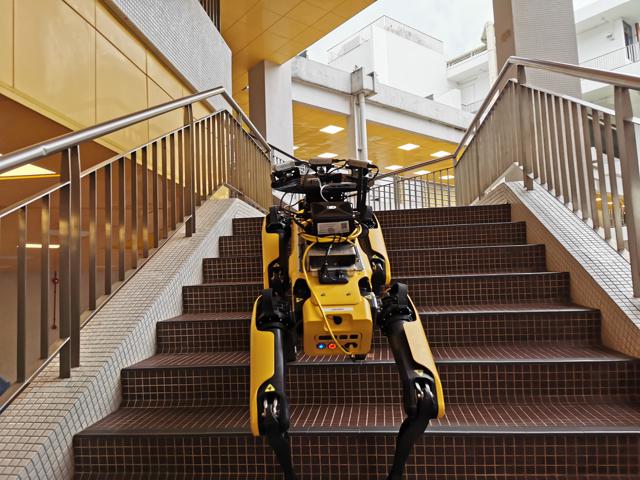} &
    \includegraphics[width=0.4\columnwidth]{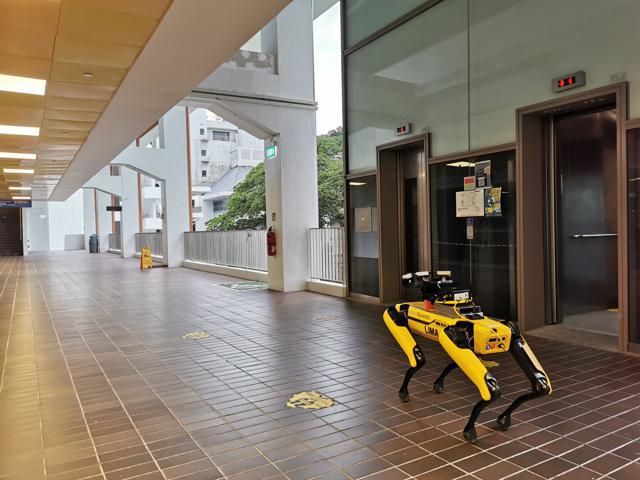} &
    \includegraphics[width=0.4\columnwidth]{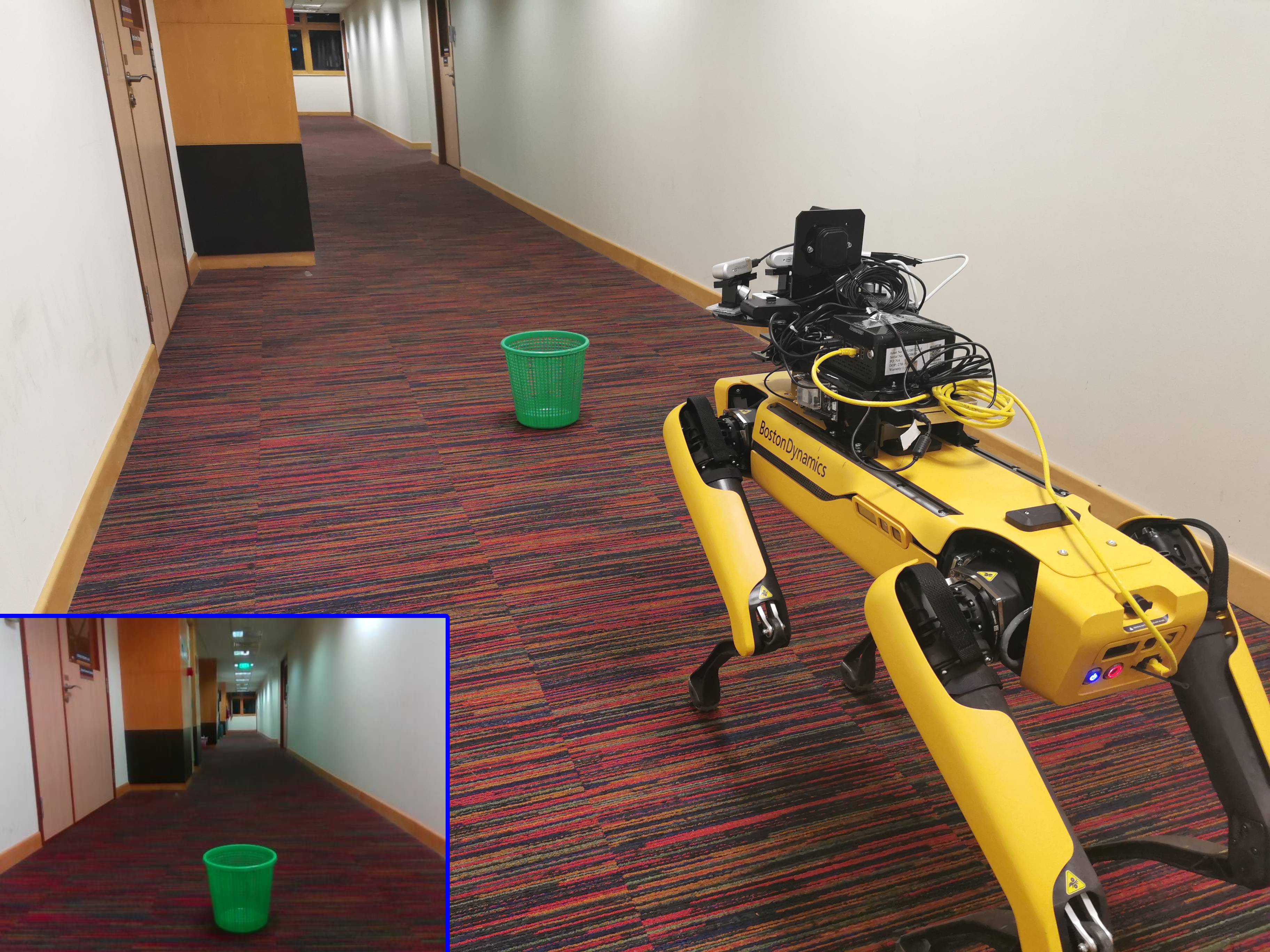} &
    \includegraphics[width=0.4\columnwidth]{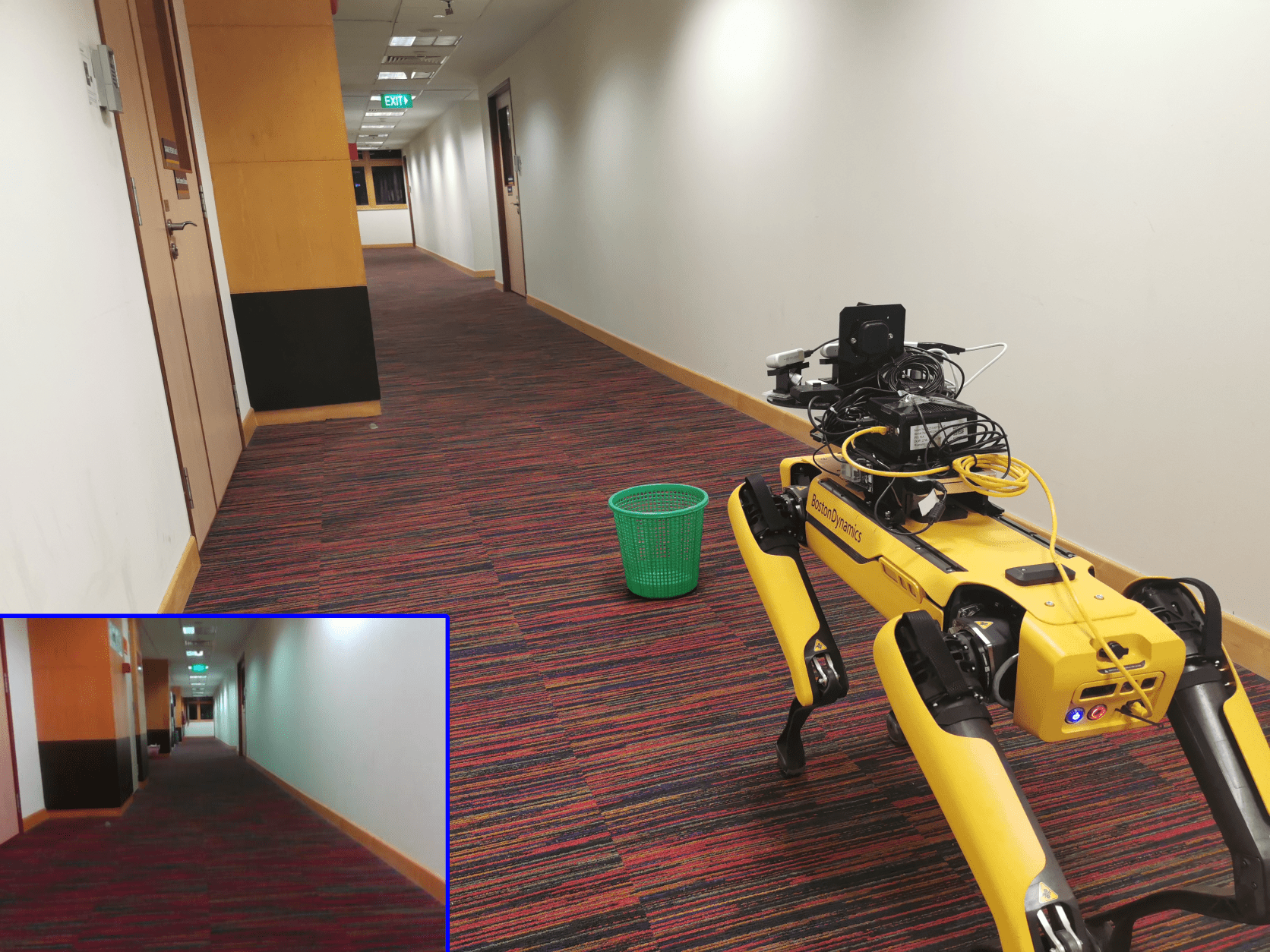}  &
    \includegraphics[width=0.4\columnwidth]{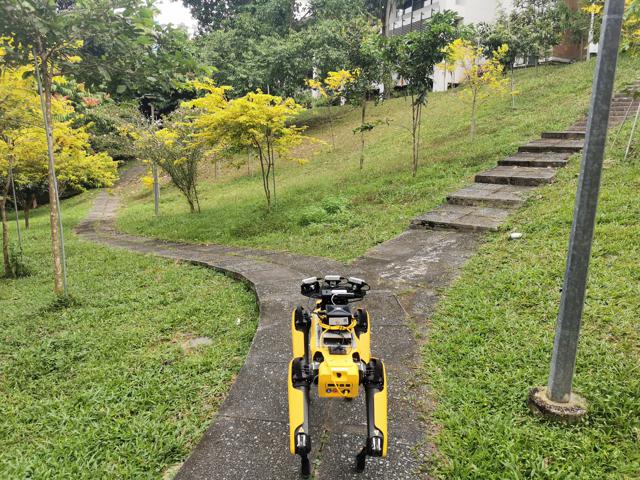} \\
    (\subfig{a}) & (\subfig{b}) & (\subfig{c}) &  (\subfig{d}) & (\subfig{e}) \\
  \end{tabular} 
  \vspace{-10pt}
  \caption{Our learned visual navigation controller running on the Spot robot. (\subfig{a}--\subfig{e}) The robot navigates in rich, diverse environments, both indoors and outdoors. (\subfig{c}--\subfig{d}) The wastebasket basket ``disappears'' as the robot gets close, because the camera has a limited field of view. The robot relies on the observation history to avoid stepping on the basket. The inset shows the view from the robot camera. (\subfig e) At a junction, the robot control depends on both the local visual perception and the global navigation objective.}
  \label{challenges}
  \vspace{-10pt}
\end{figure}

}]





\blfootnote{The authors are with the School of Computing, National University of Singapore, Singapore 117417, Singapore.}
\blfootnote{Our demo video and code are available at: \video{}.}


\begin{abstract}

How can a robot navigate successfully in rich and diverse environments, indoors or outdoors, along office corridors or trails on the grassland, on the flat ground or the staircase? To this end, this work aims to address three challenges: (i) complex visual observations, (ii) partial observability of local visual sensing,  and (iii) multimodal robot behaviors conditioned on both the local environment and the global navigation objective. We propose to train a neural network (NN) controller for local navigation via imitation learning. To tackle complex visual observations, we extract multi-scale spatial representations through CNNs. To tackle partial observability, we aggregate multi-scale spatial information over time and encode it in LSTMs. To learn multimodal behaviors, we use a separate memory module for each behavior mode. Importantly, we integrate the multiple neural network modules into a unified controller that achieves robust performance for visual navigation in complex, partially observable environments. We implemented the controller on the quadrupedal Spot robot and evaluated it on three challenging tasks: adversarial pedestrian avoidance, blind-spot obstacle avoidance, and elevator riding. The experiments show that the proposed NN architecture significantly improves navigation performance. 
\end{abstract}

\section{INTRODUCTION}



How do humans navigate? We not only navigate in a rich, diverse world with almost exclusively visual sensing, but also compose different behaviors to reach a destination: following paths, making turns, climbing staircases, avoiding obstacles, and using tools, \eg, elevators and buses. This is not yet possible for robots. Classic model-based robot controllers are susceptible to sensor noise and modeling errors. The recent data-driven neural network controllers are more robust because of end-to-end learning from data, but existing works are limited to restricted environments, \eg, driving on mostly empty lanes \cite{nvidia_il, il_racing}. We seek to scale this approach to more realistic settings where many challenges are present, as illustrated in \figref{challenges}.

To achieve this, we would like the robot to discover useful visual cues and learn complex navigation skills from data. In this work, our objective is to design a neural network controller with the structural components to  (\romannum{1}) extract rich spatial features from visual perception, 
(\romannum{2}) capture rich history information to account for the partial observability of local sensing, and 
(\romannum{3}) generate multimodal output distributions conditioned on the local environment and the global navigation objective. 

\begin{figure*}[t]
  \centering
  \includegraphics[width=0.85\textwidth]{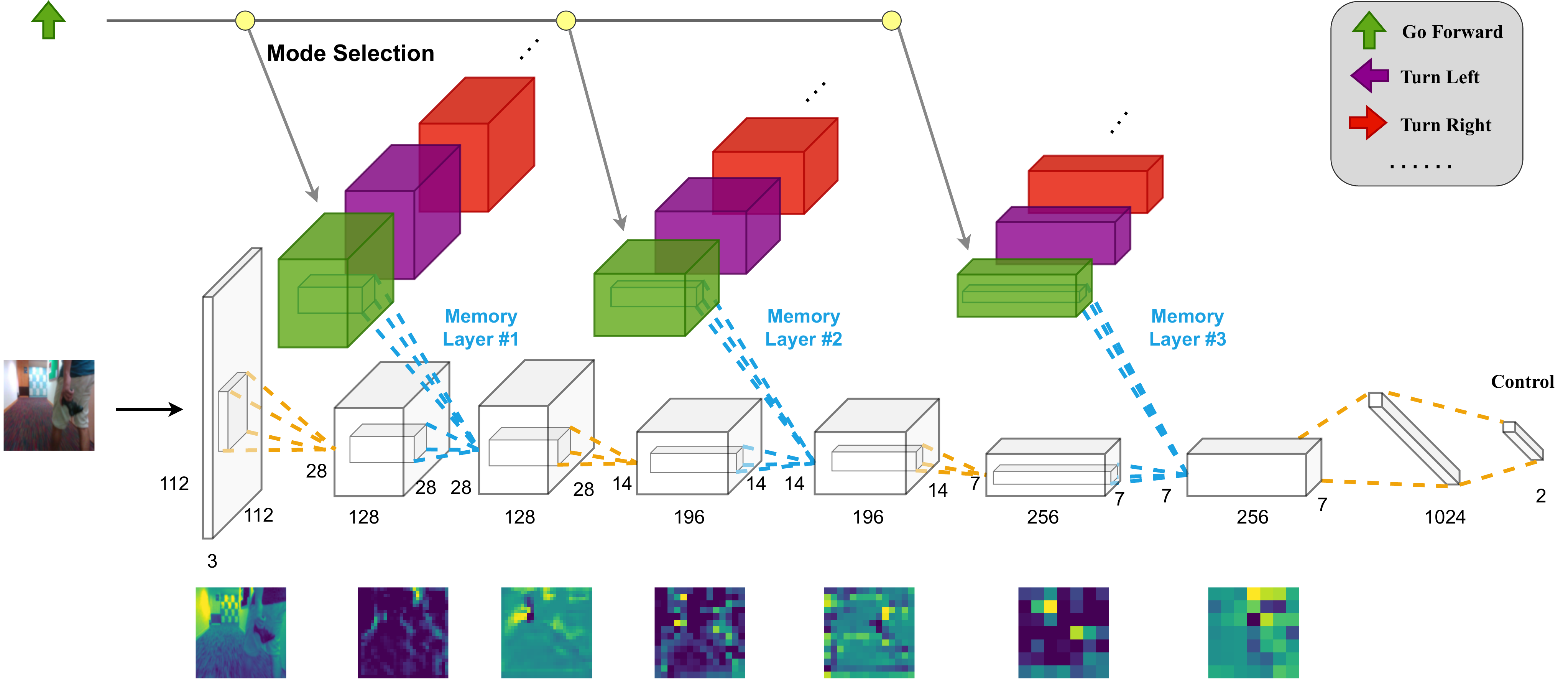} \\
  \vspace{-5pt}
  \caption{The neural network architecture of our \proposed{} controller. Every 3D volume in the figure denotes a feature map of shape (channel, width, height). The colored volumes denote the latent representation of the history, and blank volumes denote the spatial features at the current time step. Dashed lines between volumes denote operations such as convolutions. Key ideas: As features propagate through the convolutional layers, the representation becomes more abstract (visualized in bottom row), and the memory layers (\cell{1}-\cell{3}) integrate history information at multiple abstraction levels to enrich the representation. To learn multimodal behaviors, the memory modules for different modes in each memory layer are disentangled (volumes in different colors) and a symbolic signal is used to select the corresponding memory for feature propagation. }
  \label{archi}
\vspace{-16pt}
\end{figure*}

Our design is inspired by insights from related tasks. Convolutional neural networks (CNNs) show the effectiveness of progressive subsampling in extracting abstract patterns \cite{conv_lecun, alexnet, resnet}. Mixture density networks (MDNs) \cite{mdn, mdn_sketch, mdn_limitations} show how independently predicting the modes in a mixture distribution can preserve the distinction of each mode. For partial observations, previous works exploit history for information gathering, implemented by either recurrent structures in the case of learning \cite{drqn}, or explicit belief tracking in Partially Markov Decision Process (POMDP) planning \cite{pomdp_leslie, pomcp, sarsop}. However, we will show that a straightforward composition of these ideas does not tackle the challenges combined. To this end, we propose our \fullname{} (\textbf{\proposed{}}), which has two key structural components: 
\begin{itemize}
    \item \textbf{Multi-scale temporal modeling}: We conjecture that both low-level features and abstract patterns in a CNN are informative for generating the control. Thus, we perform temporal reasoning at different levels using memory modules to account for partial observations. 
    \item \textbf{Multimodal memory}: We extend the idea of MDN \cite{mdn} to temporal reasoning and propose to maintain independent memory modules for different modes.
\end{itemize}
The architecture diagram is presented in \figref{archi}. Our \proposed{} is able to learn complex policies from data and can be incorporated to different navigation architectures (\eg, \cite{inet, cil, dan, spa_temp_coh_maps}). In this work, we integrate it into the Intention-Net framework that performs global path planning on a crude map and generates intention signals to guide local maneuvering \cite{inet}. Our \proposed{} can achieve robust navigation in our university campus and has been in operation for a cumulative total of about 200 hours. \footnote{Estimated value from April 2021 to the time of writing.}

To evaluate \proposed{}, we collect human demonstration data and train the controller with supervised learning. We evaluate the multi-scale temporal representation design on tasks with highly partial observations (\secref{exp:po}), and evaluate the multimodal memory structure on tasks that require switching between different behavior modes (\secref{exp:multimodal}). Across all tasks, our model significantly outperforms alternative designs, including CNNs, LSTMs, or the ablated versions of our model.

\afterpage{}

\section{RELATED WORK}
Robot navigation systems often adopt a hierarchical architecture \cite{survey_archi_program}. The classical model-based system design consists of components for mapping and localization \cite{orbslam, orbslam2, real_mon_slam, vslam-ar}, planning \cite{rrt, exp_plan, prm}, and control \cite{pred_act_control, traj_control}. 
The premise of this architecture is to tackle the complexity of robot navigation through decomposition. 
However, these components depend on handcrafted models and may be brittle, because of modeling errors or imperfect perception.
Moreover, they require manual intervention in order to adapt to environmental changes, such as repaved roads. 
These limitations motivate recent works that use end-to-end learning from data to reduce dependency on accurate models and maps \cite{variational, cil, inet}. 

These model-free controllers learn a mapping from raw sensory data to the control end-to-end. It sidesteps the difficulty of hand-crafting models for planning, which is particularly challenging for visually rich and complex domains. The end-to-end learning of control requires supervision either from reinforcement learning (RL) or from imitation learning (IL). While RL in simulation suffers from the simulation-to-reality gap \cite{drive_a_day, rl_multiagent, dl_driving}, IL has been more successful in real-world tasks. It has been used to learn lane following \cite{alvinn, nvidia_il}, static obstacle avoidance \cite{offroad_obstacle}, and car racing \cite{il_racing}. Furthermore, imitation control policies have been incorporated into map-lite navigation systems \cite{variational, cil, inet}. However, there has been little work targeting the challenge of partial observability in visual navigation. In addition, we address the need for learning multimodal behaviors, which is essential to completing a long-horizon navigation task.

\section{Navigation with \proposed{}}

\subsection{Problem Formulation}
Our objective is to achieve goal-directed visual navigation with a local controller that has a pre-defined set of modes $M =$ \{\intentforward{}, \intentleft{}, \intentright{}, \intentelevator{}\}. At time step $t$ during navigation, the environment is in some state $s_t$. The controller $\pi$ receives a raw RGB observation $o_t$ and a mode signal $m_t \in M$ that instructs the behavior of the local controller. Next, it updates its memory $c_{t-1}$ with a module $g$ conditioned on $m_t$, \ie
$$
c_{t} = g_{m_t}(o_t, c_{t-1})
$$
The controller outputs an action $\hat{a_t}$ which is a tuple of steering and velocity $\langle \theta_t,v_t \rangle$, \ie
$$
    \hat{a_t} = \pi_{m_t, c_{t}}(o_t)
$$
The intuition is that the memory $c_t$ is a latent representation of history that estimates task-relevant aspects of the states $s_t$ from the partial observations $o_t$. Instead of hand-crafting representations, we allow the controller to discover useful patterns through learning. In this work, we adopt a supervised paradigm and optimize the $L2$ loss\footnote{However, we find that the loss value in offline training does not correlate well with the online performance, due to the mismatch between the i.i.d. assumption of the loss and the sequential nature of the task \cite{offline_eval}. For relevance to real-world tasks, we conduct online evaluations in the physical world.} over a human demonstration dataset $\mathcal{D} = \{\langle o_t, m_t, a_t \rangle\}_t$, \ie 
$$
    \mathcal{L} = \mathbb{E}_{\langle o_t, m_t, a_t \rangle \in \mathcal{D}} \|\pi_{m_t, c_{t}}(o_t) - a_t \|_2
$$

We parameterize $\pi$ with a neural network and optimize the objective end-to-end. We will show that the parameterization of $\pi$ is important: with the same data, our model learns a policy that is significantly better than alternative designs.

\subsection{\proposed{} Architecture}
In this section, we introduce the architecture of our \proposed{}. We begin with the formulation of our memory module, followed by the multi-scale history integration and the multimodal memory cell designs. 

\subsubsection{\textbf{Memory Module}}
We use a modified Convolutional LSTM (ConvLSTM) \cite{convlstm} to aggregate spatio-temporal information over time. ConvLSTM is a variant of peephole LSTM  \cite{lstm_peep} where the input and output are 2D feature maps instead of 1D feature vectors. We additionally incorporate structures that ease optimization and improve generalization. 

\textbf{Normalization layers.} We use a GroupNorm \cite{group_norm} layer ($G=32$) after each convolution in the cell. There are two practical reasons. First, we need normalization to stabilize the gradient flow, as we observe that the gradients appear statistically unstable when they propagate through deep layers and long time horizons, which makes training much harder. Second, normalization reduces variance in the intermediate features. When batch size is small, \ie, a batch size of one at deployment time, features show more statistical variability than at training time, which destabilizes runtime performance. Among different normalizations \cite{layer_norm, group_norm, batch_norm}, we find GroupNorm \cite{group_norm} with a large $G$ the most effective.

\textbf{Dropout.} As the observations in adjacent time steps are highly correlated, we need to prevent the memory module from overfitting to the redundancy in the temporal features. We adapt 1D dropout methods for RNNs \cite{rnn_drop, recurrent_dropout, mc_dropout} to 2D features by applying it channel-wise. We find that jointly using recurrent dropout \cite{recurrent_dropout} and Monte Carlo dropout \cite{mc_dropout} yields the best real-world performance. 

\textbf{Overall structure.} Firstly, dropout is applied to the input $x_t$ and hidden state $h_{t-1}$ \cite{mc_dropout} 
\begin{align*}
    x_t = d(x_t),\:h_{t-1} = d(h_{t-1})
\end{align*}

Then, we compute the three gates $i_t$, $f_t$, and $g_t$, 
\begin{align*}
     i_t & = \sigma(GN(W_{xi} \circ x_t + W_{hi} \circ h_{t-1} + W_{ci} \circ c_{t-1} + b_i)) \\
     f_t & = \sigma(GN(W_{xf} \circ x_t + W_{hf} \circ h_{t-1} + W_{cf} \circ c_{t-1} + b_f)) \\
     g_t & = tanh(GN(W_{xc} \circ x_t + W_{hc} \circ h_{t-1} + b_c))
\end{align*}

In addition, dropout is applied to the cell update $g_t$ to alleviate overfitting to redundant temporal cues \cite{recurrent_dropout}
$$
    g_t = d(g_t)
$$
$$
    c_t = f_t * c_{t-1} + i_t * g_t  \vspace{-8pt}
$$
Finally, we update and output the hidden state  
$$
    o_t = \sigma(GN(W_{xo} \circ x_t + W_{ho} \circ h_{t-1} + W_{co} \circ c_t + b_o)) 
$$
$$
    h_t = o_t * tanh(c_t) \vspace{-8pt}
$$
where $d$ is the dropout operation, $W$ represents model weights, $\circ$ denotes convolution, $*$ is the Hadamard product, and $GN$ denotes GroupNorm. In this way, history is encoded recursively into $c_t$ across time steps, which is used to generate an enriched representation $h_t$ from the input $x_t$.

\subsubsection{\textbf{Multi-Scale Temporal Modeling}} 
With the memory cell, we propose to aggregate history information at multiple abstraction levels to account for partial observability. This is achieved by inserting multiple memory layers into a CNN.

\textbf{Motivations from the angle of perception.} In the early layers of a CNN, high-resolution feature maps contain low-level coarse features, \eg, edges and textures, which are useful for pixel-level prediction tasks (\eg, \cite{convlstm, fcn, unet}); in the late layers, the features encode abstract patterns of the visual input, \eg, objects and shapes, which are critical to recognition (\eg, \cite{fpn, s3d}). For navigation tasks, we argue that temporal reasoning jointly on both features yields optimal performance. One example of low-level temporal cues is that, to avoid a moving obstacle, the agent needs to forecast its future positions, and it has been shown that motions can be more accurately captured from high-resolution features maps \cite{convlstm}. Furthermore, high-level temporal understanding is critical to recognizing environment dynamics, \eg, distinguishing door opening from door closing. To have the best of the two worlds, we extract temporal information at multiple spatial scales. 

\textbf{Tightly coupled perception and control.} On the other hand, since \proposed{} is trained end-to-end, the perception and control are tightly coupled. Thus, beyond capturing perceptual cues of varying granularities, different layers may also learn different behavioral abstractions. This is partially supported by our observation that separating memory modules helps learn multimodal behaviors (\secref{exp:multimodal}). This also motivates our multi-scale network design. 

\textbf{Implementation.} We insert memory layers into a CNN (\figref{archi}). As features propagate through the model, the representation becomes more abstract and the memory layers (\cell{1}-\cell{3}) fuse information from the past and the current time step to enrich the representation. In \secref{exp:po}, we show that this multi-level integration can better compensate for partial observations and yield significantly more robust controls, compared to memory at a single level.

\subsubsection{\textbf{Multimodal Memory Structure}} \label{mmc:multimodal} 
For a single observation, the agent may need to perform different actions in order to follow the high-level navigation objective. Thus, it is important to learn a conditional multimodal policy. The problem here is mode collapse, \ie, it may learn a unimodal policy that is averaged from different behaviors. 

Our solution is to disentangle representations for different modes. This is intuitive: For example, the modes \intentforward{} and \intentleft{} require different actions from the controller and lead to distinct dynamics in the observation. One temporal modeling module may not be able to learn the difference from data efficiently. Thus, we enforce such distinction by incorporating explicit structures. Specifically, there are multiple memory cells in one memory layer; for each input, only one cell is activated to propagate features (\figref{archi}). The selection is instructed by a mode signal generated by high-level planning. We will show the effectiveness of the design in \secref{exp:multimodal}.

\subsection{Data Collection}
We implement the full pipeline from data collection to deployment on a Boston Dynamics Spot robot. The RGB signals come from an Intel RealSense D435i camera that is connected to an NVIDIA AGX Xaiver onboard computer.

\begin{figure}[h]
  \centering
  \setlength{\tabcolsep}{1pt} 
    \renewcommand{\arraystretch}{1} 
    \vspace{-3pt}
  \begin{tabular}{ccc}
    \includegraphics[width=0.32\columnwidth]{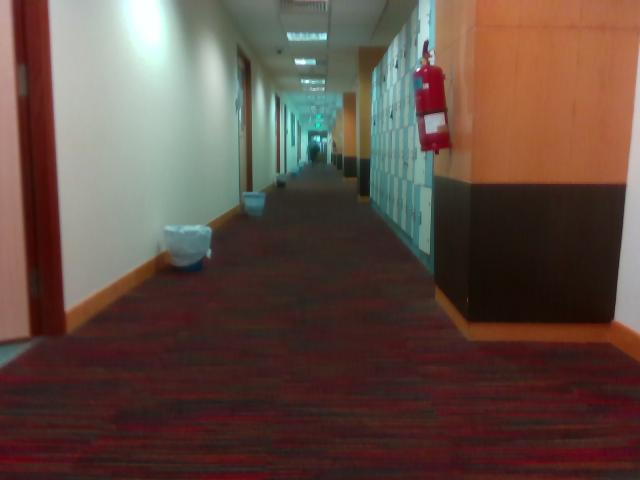} &
    \includegraphics[width=0.32\columnwidth]{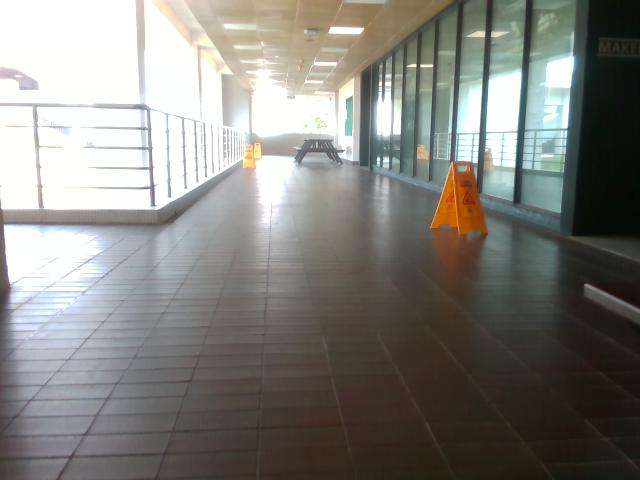} & 
    \includegraphics[width=0.32\columnwidth]{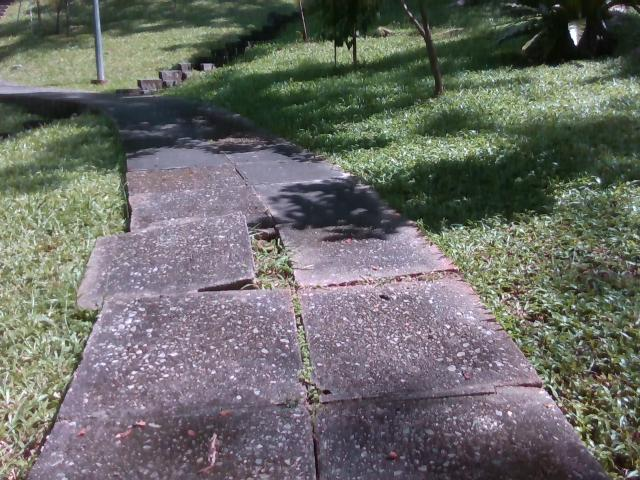} \cr
  \end{tabular}
  \vspace{-5pt}
  \caption{Sample images from the dataset: (from left to right) indoor, semi-outdoor, and outdoor environments. Note that sensor noise, \eg, glaring, and environment noise, \eg, shadows, are prevalent in the physical world.}
  \label{data_collection_env}
  \vspace{-13pt}
\end{figure}

We collect expert demonstrations by teleoperation. The dataset has 2 parts. In the first part, we navigate various environments (\figref{data_collection_env}), where there are natural or a small number of manually placed obstacles, \eg, pedestrians and bulky objects. In the second part, we collect data specifically for \taskref{elevator} with HG-DAGGER \cite{hg_dagger}, because the task is highly sequential and prone to errors. In total, our dataset contains 410K time steps, or equivalently 7.6 hours of recording, in which 64K steps are from the second part.

\textbf{Data distribution matters.} A crucial step after the collection is to carefully moderate the dataset distribution. The objective is to balance different components in the dataset to prevent the model from only learning the dominant behavior. For example, without proper dataset resampling, the controller may only learn the dominant mode \intentforward{} and never make turns or recover from deviations.

\subsection{Training}
We train our \proposed{} with Truncated Backpropagation Through Time (TBPTT) \cite{tbptt} and we find that AdamW optimizer \cite{adamw} accelerates convergence by a clear margin. We take every third frame in the dataset to construct model input sequences of length $L=35$. The data is then normalized with ImageNet \cite{imagenet} statistics and augmented with strong color jittering. In each training iteration, the model predicts controls for $k_1=5$ observations, and backpropagation is done every $k_2=10$ predictions. In particular, as \taskref{elevator} is a long-horizon task, we need to triple the values for $L$, $k_1$, and $k_2$ for the mode \intentelevator{}.

Analogical to the linear scaling rule \cite{imagenet_1hr, generalization_sgd}, we define our learning rate $LR$ as linear in batch size $BS$ and $k_2$, \ie, $LR = BaseLR * BS * k_2$, where $BaseLR$ and $BS$ are set to 1e-7 and 36 respectively. We use a weight decay rate of 5e-4 and a dropout rate of 0.3. Input images are resized to 112$\times$112 at both training and deployment time for fast computation. We train the model for 200 epochs and decay the learning rate at the 70th and 140th epochs. Training takes 40 hours with 4 $\times$ RTX2080Ti.

\section{Experiments}
\label{experiments}
In this section, we verify the two hypotheses that form the basis of our architecture design:
\begin{enumerate}[label=\roman*.]
    \item Temporal modeling at multiple spatial abstraction levels can better improve robustness to partial observations, compared to at a single level or no history
    \item Separating memory modules for different modes is critical to learning a multimodal policy 
\end{enumerate}

As the hypotheses focus on orthogonal aspects, we conduct independent testings (\secref{exp:po}, \secref{exp:multimodal}). For consistency, we use the same evaluation metrics across all tasks. Each task is defined to have $n$ steps and success is defined as completing all steps error-free. We repeat each task $N$ times and record the number of successful steps in each trial $s_i, i \in [1, N]$. We compute 2 metrics, success rate (\SR{}) and completion rate (\CR{}):
\begin{align*}
 \SR{} = \frac{1}{N}\sum_{i=1}^{N}\mathds{1}_{[s_i=n]} \quad
 \CR{} = \frac{1}{N}\sum_{i=1}^{N}\frac{s_i}{n} 
\end{align*}
\SR{} is the prime performance indicator that estimates the likelihood of completing a task, whereas \CR{} is a relaxed measure that is more granular to differentiate policies.

In each of the following subsections, we first introduce tasks and baselines, then analyze experiment results.

\subsection{Navigation Under Partial Observability}
\label{exp:po} This subsection answers the question: Does integrating memory at different abstraction levels confers advantages?

\begin{figure}[t]
  \vspace{3pt}
  \centering
  \setlength{\tabcolsep}{0.5pt} 
  \renewcommand{\arraystretch}{1} 
  \begin{tabular}{ccccc}
    (\subfig{a})
    & 
    \raisebox{-.5\height}{\includegraphics[width=0.11\textwidth]{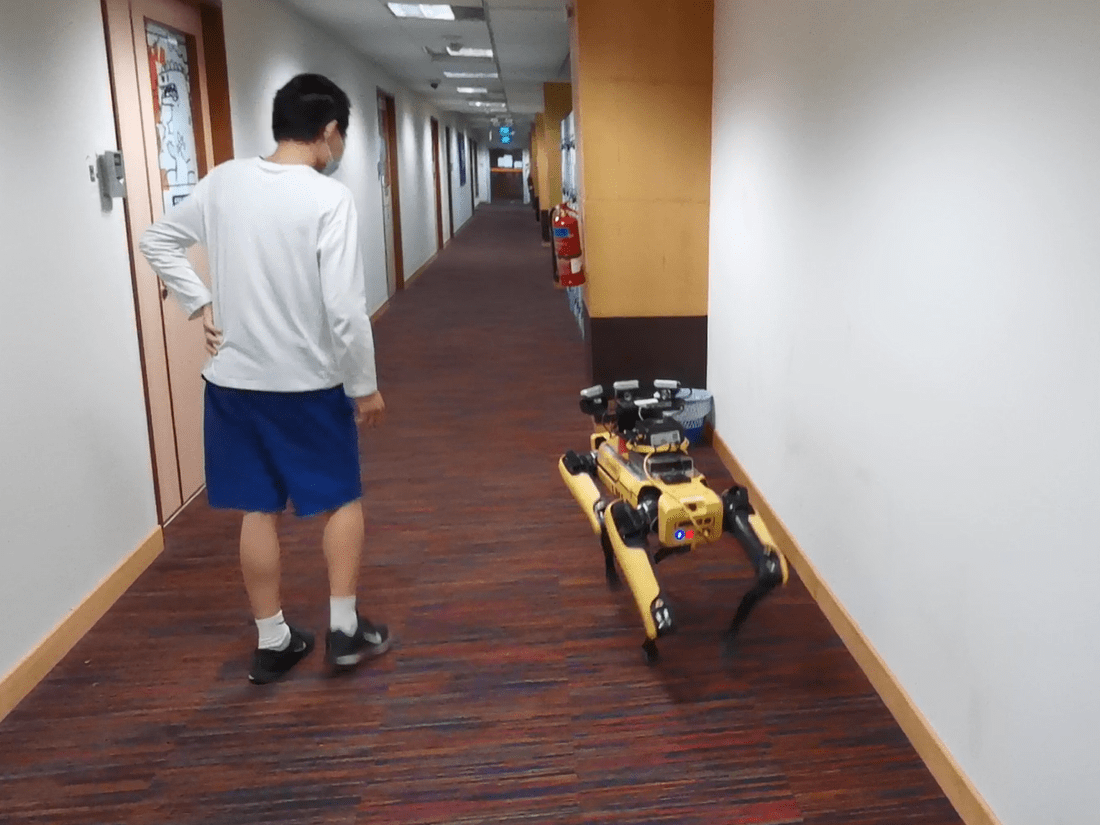}} &
    \raisebox{-.5\height}{\includegraphics[width=0.11\textwidth]{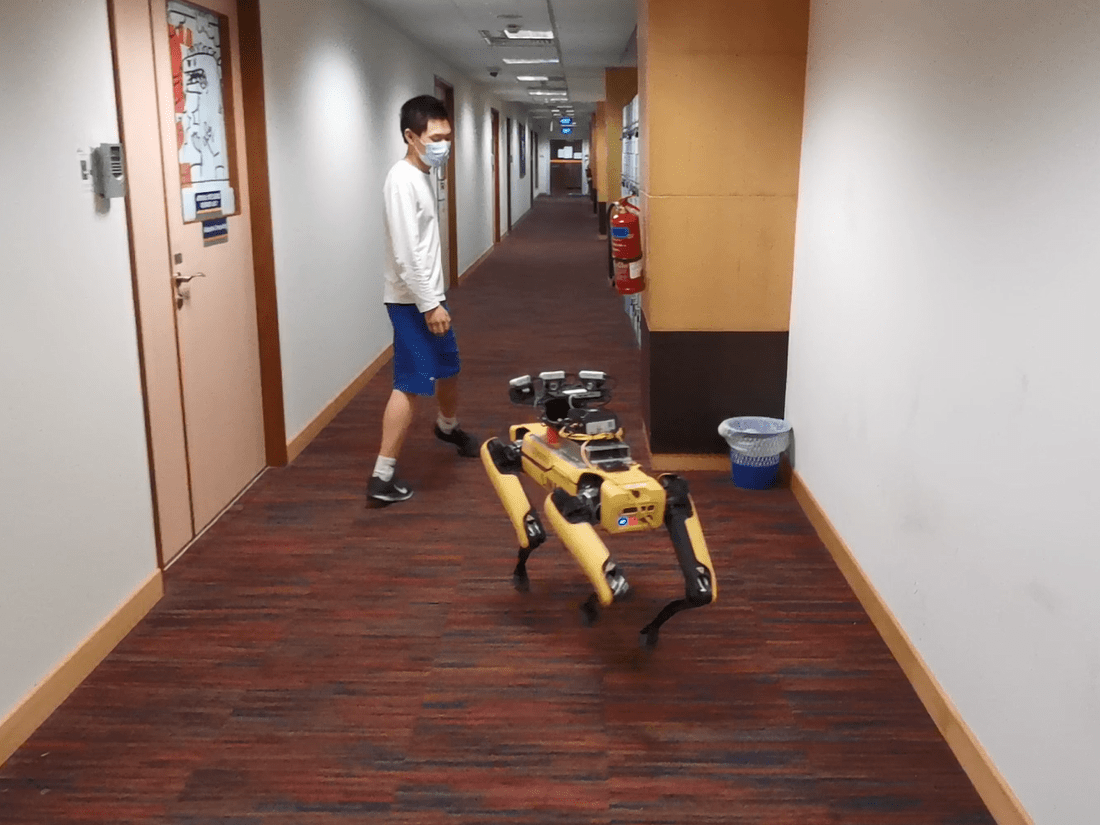}} &
    \raisebox{-.5\height}{\includegraphics[width=0.11\textwidth]{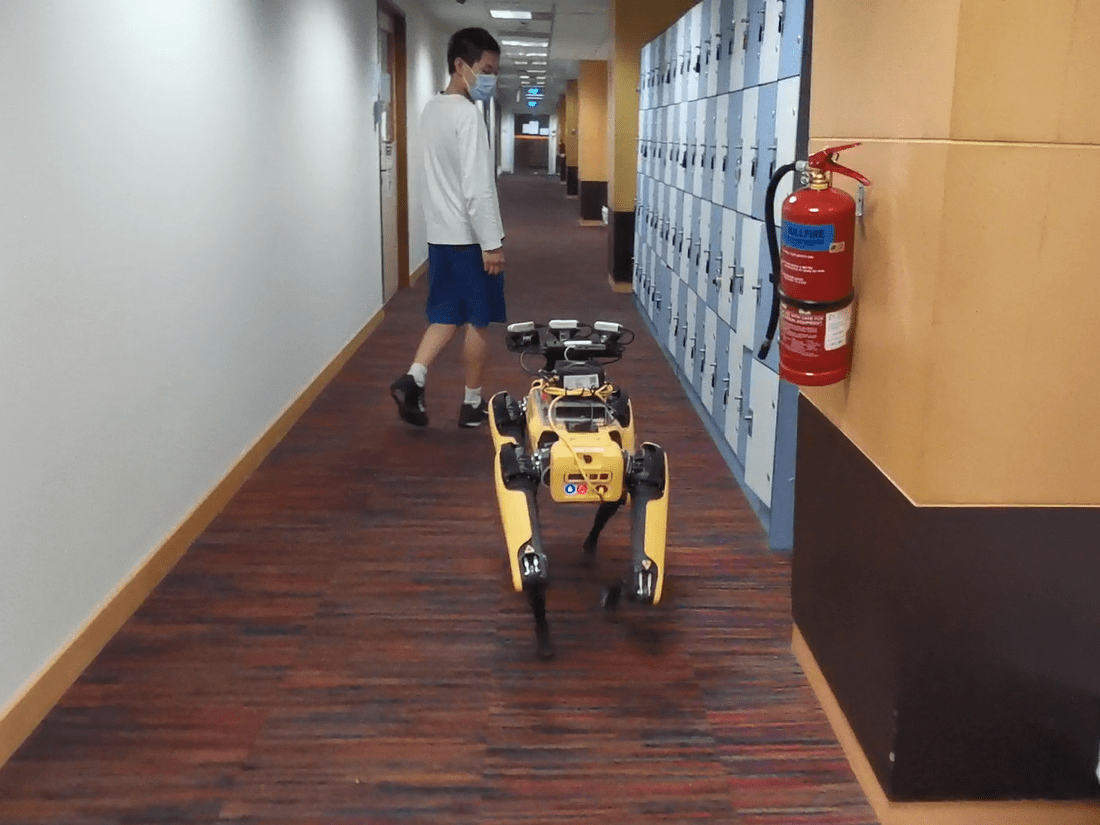}} &
    \raisebox{-.5\height}{\includegraphics[width=0.11\textwidth]{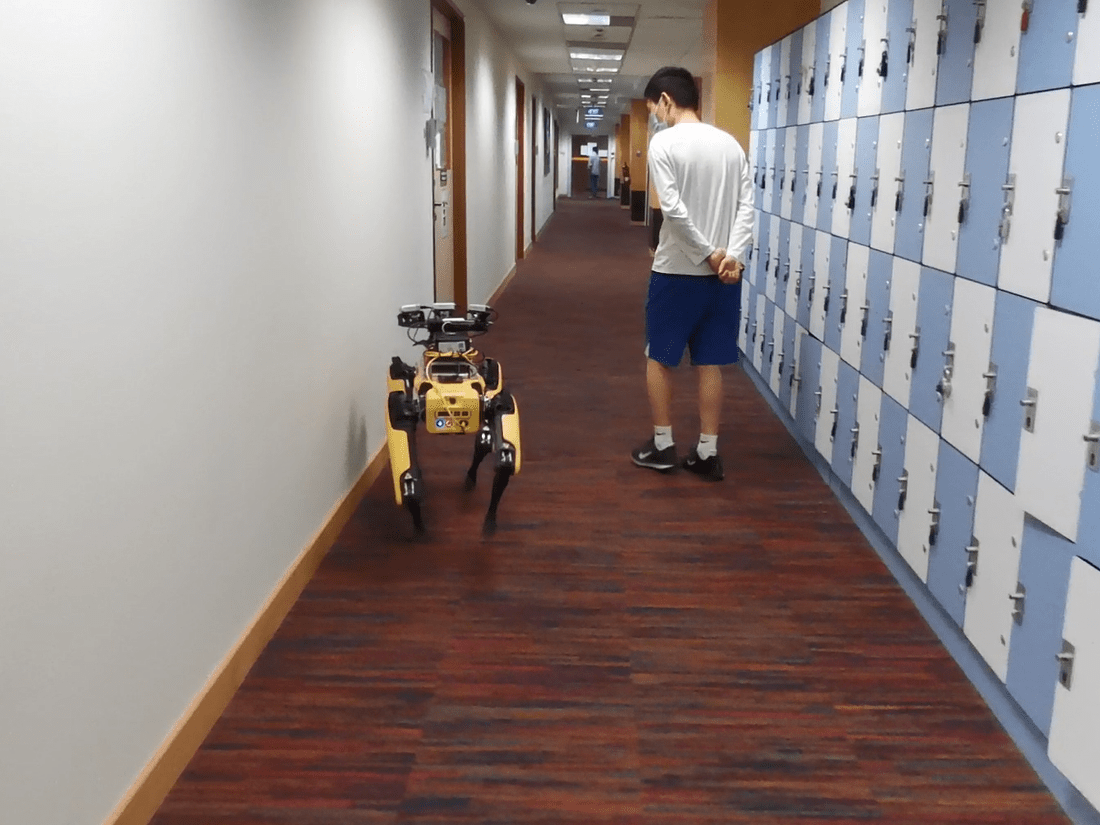}}
    \vspace{1pt}
    \cr
    (\subfig{b}) & 
        \raisebox{-.5\height}{\includegraphics[width=0.11\textwidth]{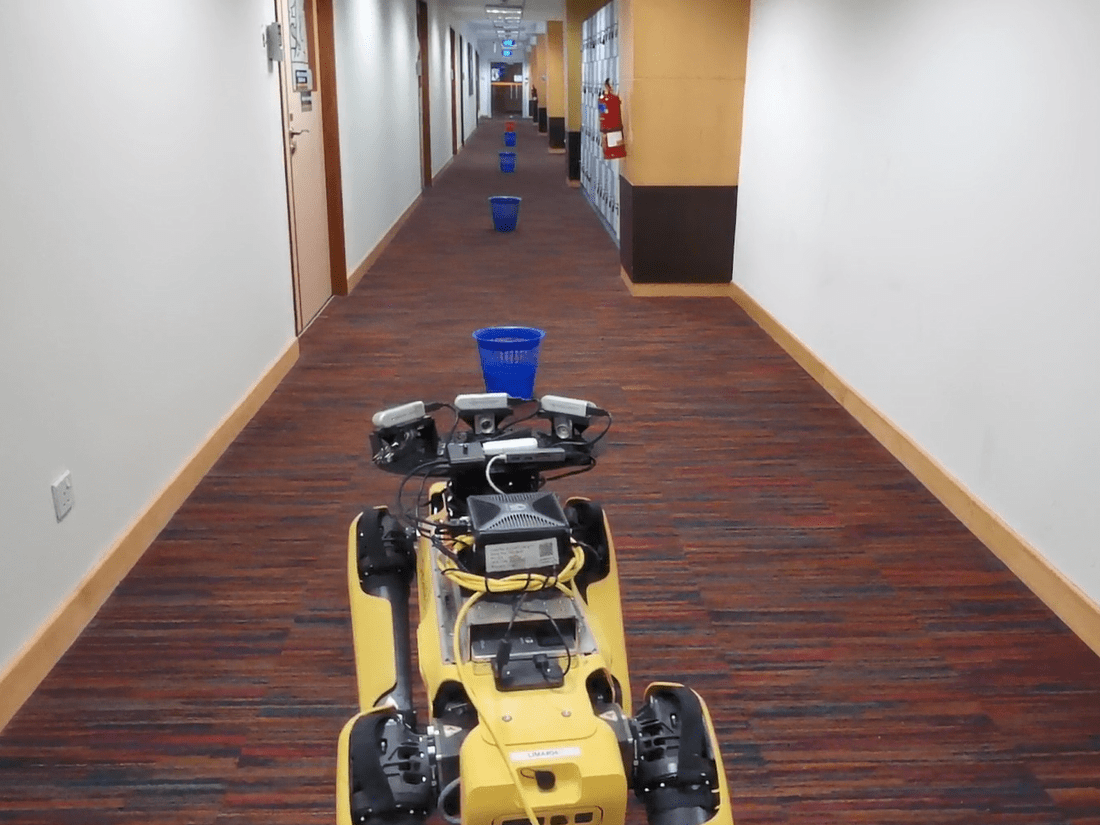}} &
    \raisebox{-.5\height}{\includegraphics[width=0.11\textwidth]{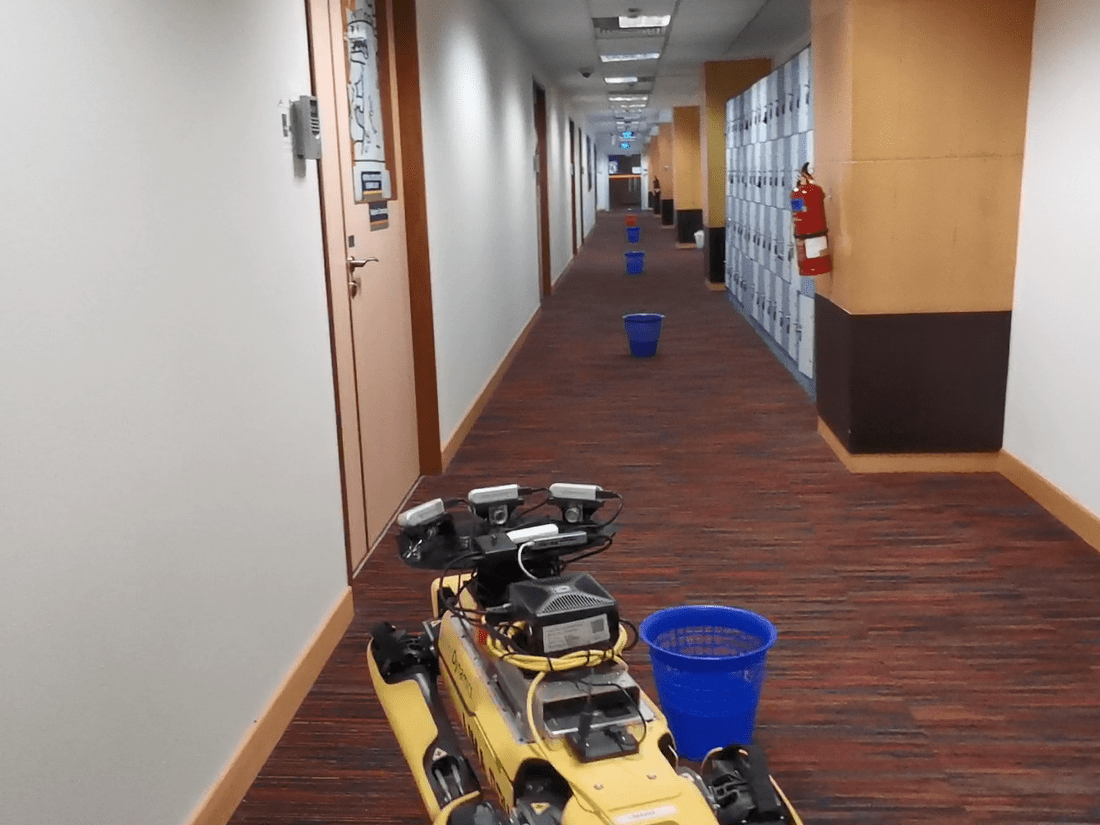}} &
    \raisebox{-.5\height}{\includegraphics[width=0.11\textwidth]{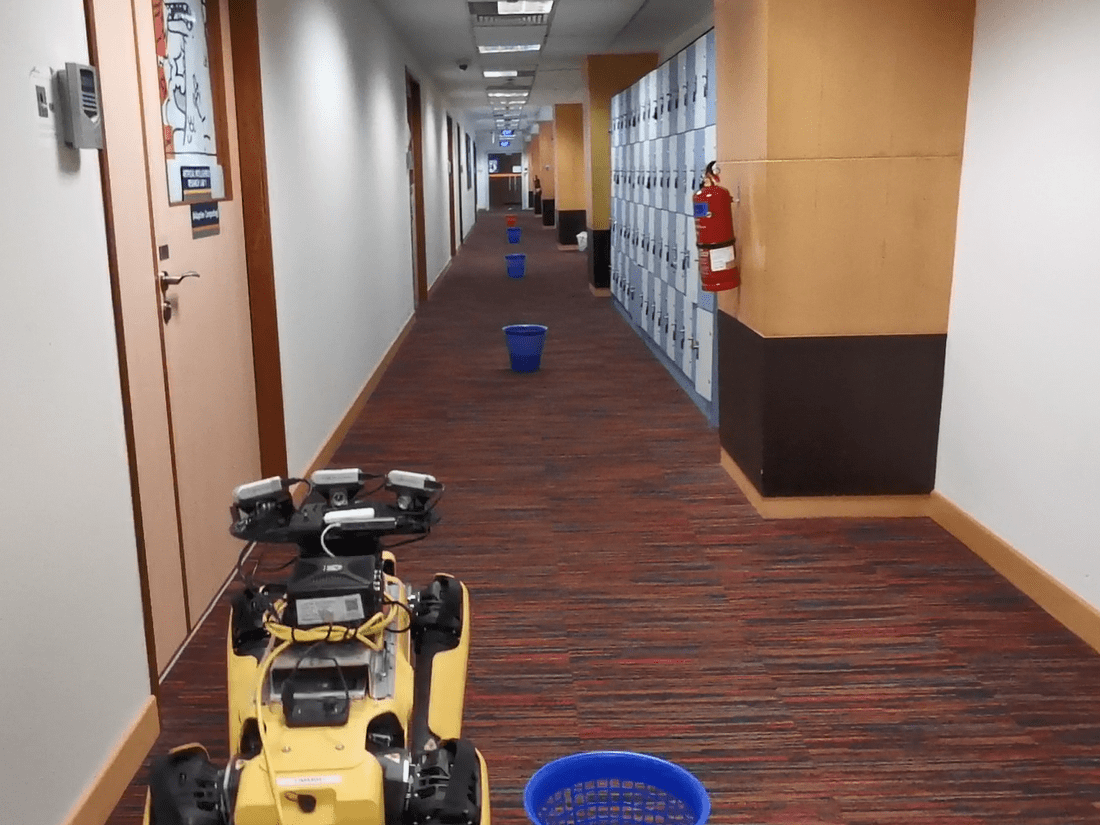}} &
    \raisebox{-.5\height}{\includegraphics[width=0.11\textwidth]{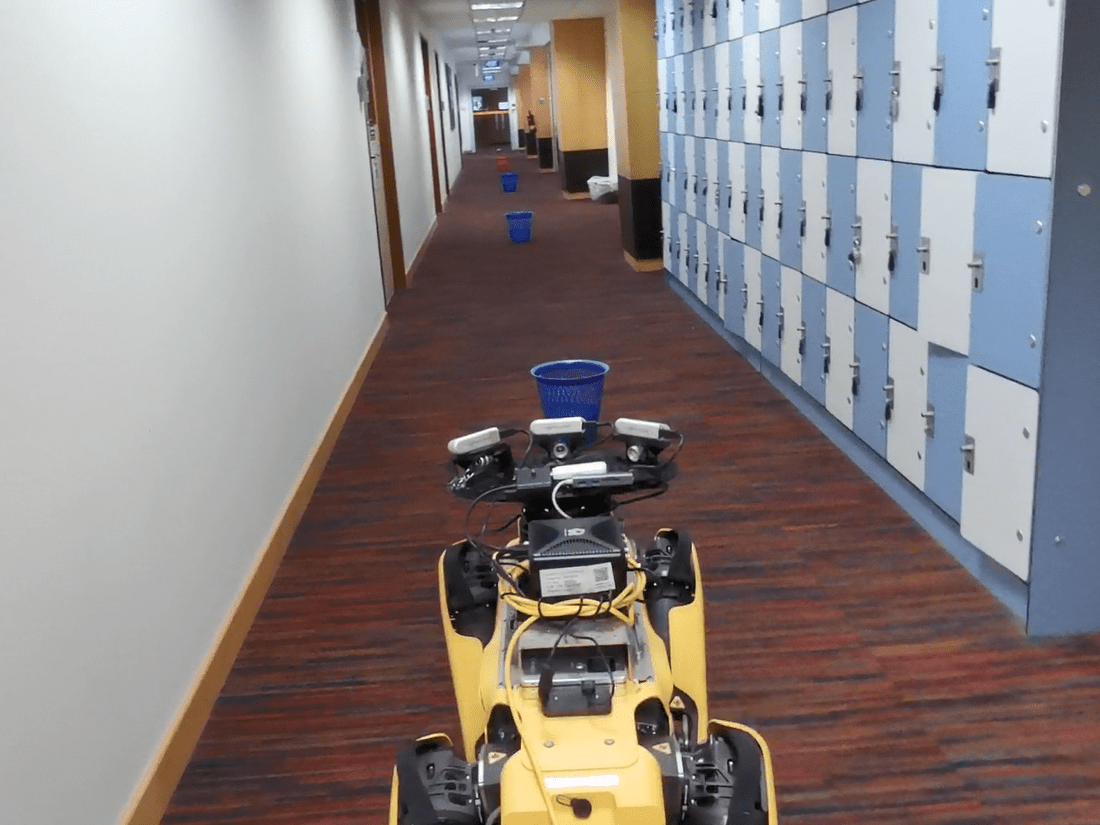}}
    \vspace{1pt}
    \cr
    (\subfig{c}) & 
    \raisebox{-.5\height}{\includegraphics[width=0.11\textwidth]{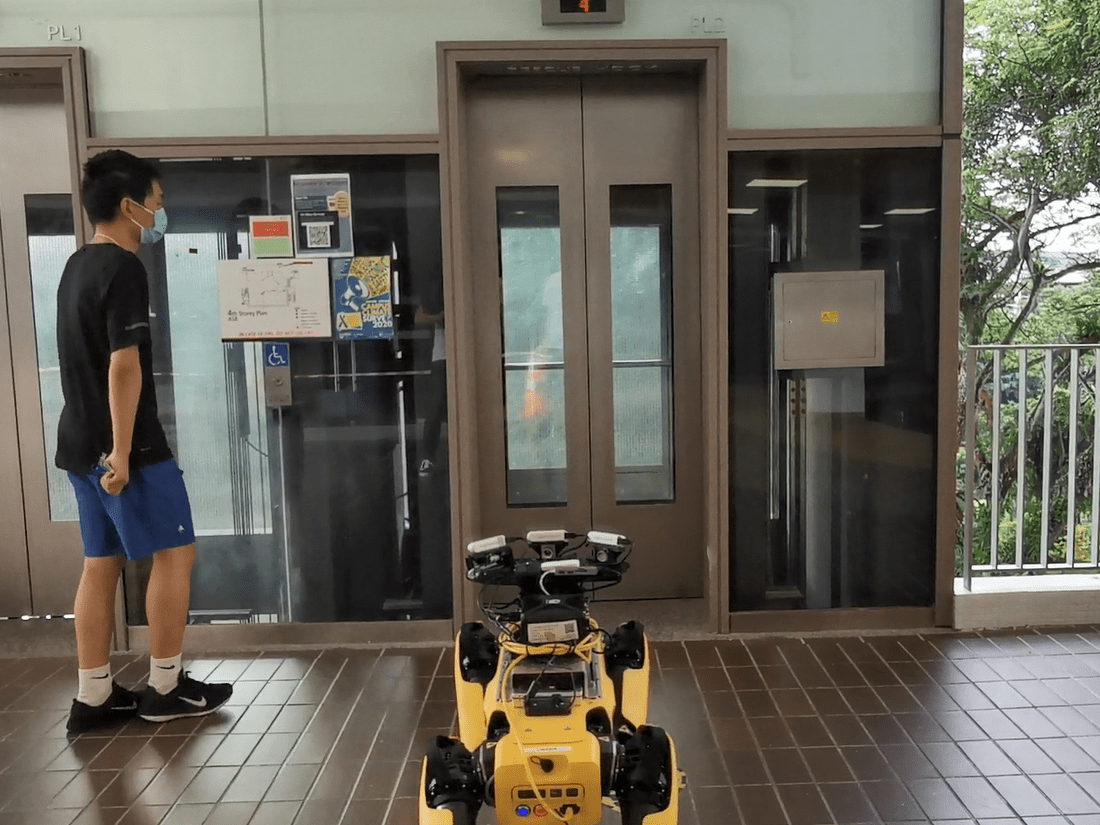}} &
    \raisebox{-.5\height}{\includegraphics[width=0.11\textwidth]{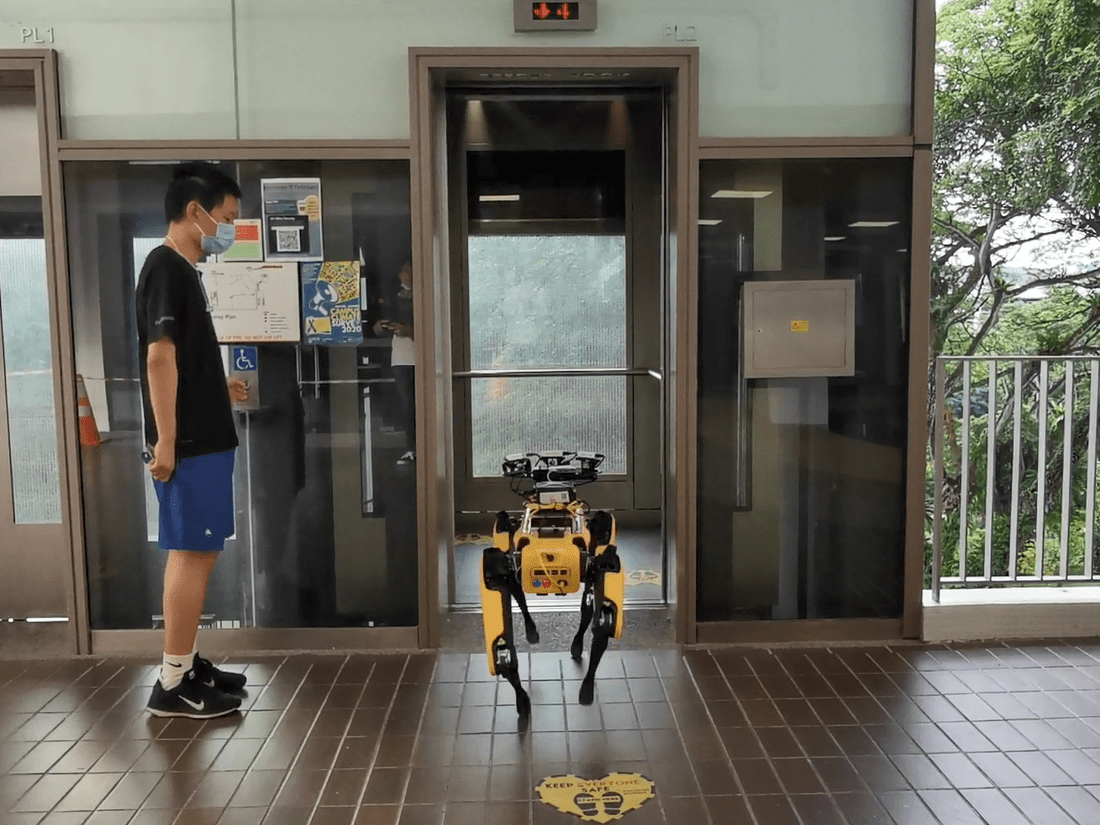}} &
    \raisebox{-.5\height}{\includegraphics[width=0.11\textwidth]{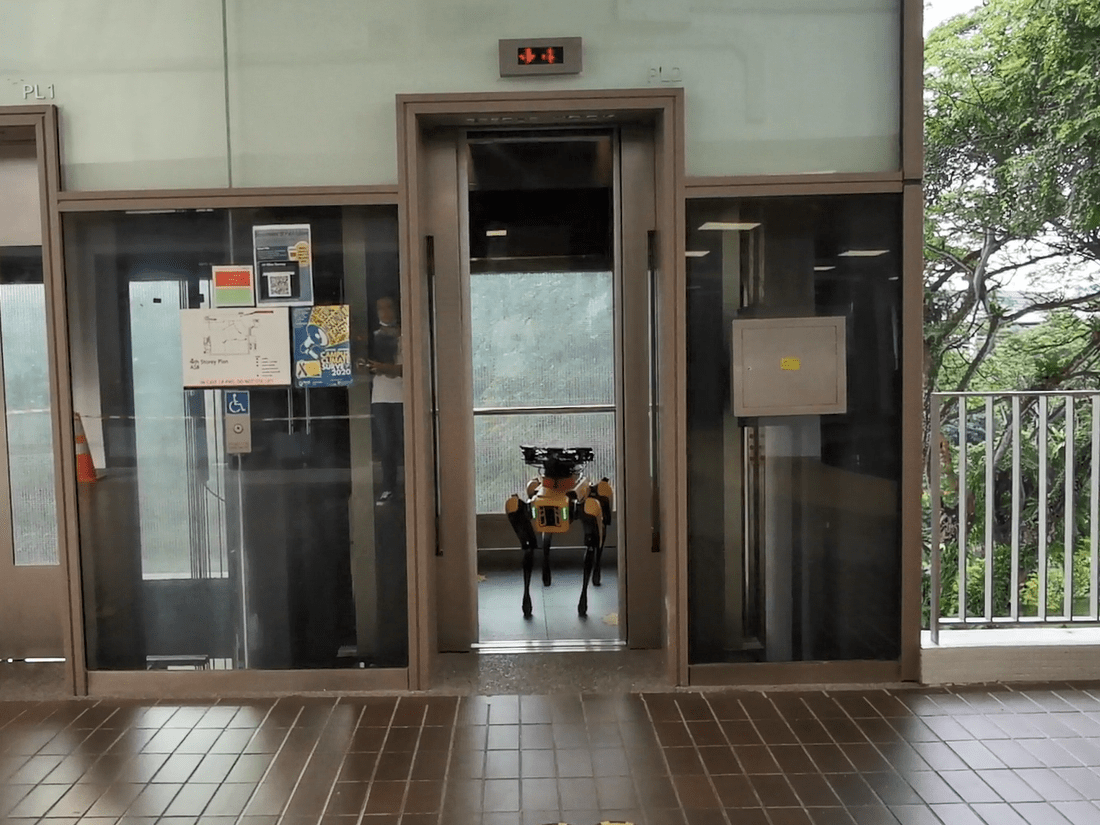}} &
    \raisebox{-.5\height}{\includegraphics[width=0.11\textwidth]{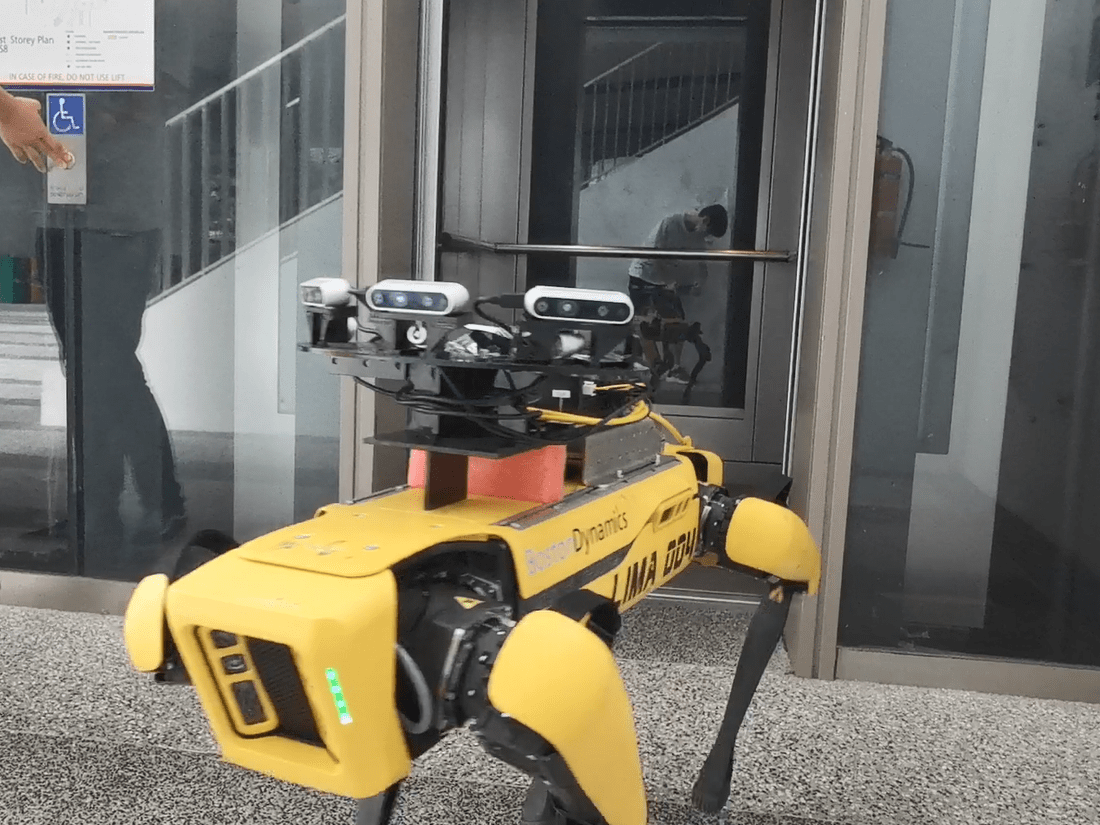}} \cr
  \end{tabular} 
  \caption{Illustrations of tasks \ref{pedestrian} - \ref{elevator}. (\subfig{a}) \taskref{pedestrian}: The robot avoids the adversary by predicting its movement. (\subfig{b}) \taskref{trashbin}: The robot successfully bypasses the basket that is not in the robot camera view. (\subfig{c}) \taskref{elevator}: The robot waits for the elevator door to open, enters the door, turns around, waits for the door to close, and exits at the other level. }
  \label{tasks}
  \vspace{-17pt}
\end{figure}

\subsubsection{\textbf{Tasks}}
We design three challenging tasks (\figref{tasks}) in which observations are highly partial:
\begin{enumerate}[label=\roman*.]
    \item \textbf{Adversarial pedestrian avoidance.} In this task, the robot aims to navigate from one end of a corridor to the other end collision-free, while an adversary constantly blocks the robot. The challenge is to predict the movement of the adversary without observing its intention. The task environment is a corridor 20m long and 2.5m wide. A collision is defined as failing to maintain a safe distance of 20cm. To block the robot, an adversary takes discrete moves to blocking positions, defined as the location 1-1.2m in the heading direction of the robot. The adversary takes the next move if the robot adjusts its direction correctly or if a collision is counted. This task contains $n \approx 15$ steps and each step is to avoid the adversary once. In case of collision, the human intervenes and the test continues. \label{pedestrian}
    \item \textbf{Blind-spot object avoidance.} \label{trashbin} The sensors' limited sensing range unavoidably causes partial observability. One example is that our robot camera can no longer observe a 20cm-tall basket if it is less than 60cm away. We place 5 baskets 5m apart on the same corridor as in \taskref{pedestrian}. The agent's objective is to navigate to the other end of the corridor collision-free. Once a collision occurs, the human intervenes and the test resumes at the next basket. The task consists of $n = 5$ steps corresponding to avoiding the 5 baskets. The difficulty is to reason about the location of the obstacle that is in the camera blind spot. 
    \item \textbf{Elevator riding.} Humans are able to use tools to accomplish tasks, but to do this, we need to keep track of task progress that is not often observable. We study the scenario in which the robot wants to go from one floor to the other by taking an elevator. Specifically, the objective is to complete the following $n = 4$ steps:
        \begin{enumerate}   
            \item wait for the door to open (5-10s) and subsequently enter the door collision-free 
            \item turn around to face the elevator door in 10s
            \item wait for the door to close without exiting (5-10s) \label{step3}
            \item when the door opens again, exit the elevator in 10s  \label{step4}
        \end{enumerate}
    The biggest challenge is to distinguish between step c) and step d) because, for the same observation, \ie, an open door, the agent needs to act differently. This requires an estimation of the task progress, \ie, the floor level, based on the history. The task is also the most challenging one, because it is highly sequential: a failure in one step takes the agent to an out-of-domain state and the agent will fail all subsequent steps. In this case, one test directly terminates if any failure occurs.  \label{elevator}
\end{enumerate}
Note that no data is collected specifically for \taskref{pedestrian} or \taskref{trashbin}. For \taskref{elevator}, data is collected during the day and we test the learned policies at night. Hence, our results also imply the generalizability of the learned policies. 

\subsubsection{\textbf{Baselines}}
We categorize architectures in previous work into 3 classes and use one representative model for each class as a baseline. In addition, we vary the designs of our controller as an ablation study. Training hyperparameter search is performed individually for each model. 
\begin{enumerate}[label=\roman*.]
    \item CNN. Following \cite{inet}, we use ResNet-50 \cite{resnet} as the backbone, appended with a branched MLP. This represents reactive policies without history \cite{nvidia_il, cil, inet}. 
    \item Multi-frame CNN (\mfinet{}). A common practice to incorporate history is to concatenate the 1D features of multiple observations \cite{drqn}. To balance computation cost and performance, we use five frames as input. It represents the policies with short-term memory. 
    \item \cnnlstm{}. The model is often used to capture temporal dependencies in visual observations \cite{drqn, rl_navigation, fcn_lstm, e2e_route_planner}. It represents the policies with long-term memory of visual semantics. For a fair comparison, we remove the three memory layers in \proposed{} and insert a three-layer LSTM before the MLP.
    \item \proposed{} variants. To study the effects of the abstraction level of temporal reasoning, we obtain \proposed{} variants by removing all memory layers (\Backbone{}), keeping only the first or the third layer (\EarlyLayer{}, \LateLayer{}). The difference between \Backbone{} and \inet{} is that the former is a shallow 3-layer CNN, whereas \inet{} has a pretrained ResNet-50 as the feature extractor.
\end{enumerate}
Moreover, we wish to know the gap between the policy learned via imitation and human driving skills. To do this, we manually drive the robot to perform the tasks using the video feed from the onboard camera, \ie, the same visual input as the other controllers. 

\subsubsection{\textbf{Results}}
Full results are summarized in \tabref{performance:po}. Notably, when evaluating at the task level (\SR{}), none of the alternative models achieves sufficiently good performance, but a high \SR{} is critical for long-horizon navigation. Our \proposed{} achieves performance close to human operators, showing its capability of learning from demonstrations.

\begin{table}[t]
\centering
\scriptsize
\caption{performance under partial observability ($N=10$). Ablation study of our method removes selected memory layers}
  \label{performance:po}
  \vspace{-7pt}
 \begin{tabular}{l c c c c c c c}
    \toprule
    Model &  
    \multicolumn{2}{c}{\taskref{pedestrian}} & 
    \multicolumn{2}{c}{\taskref{trashbin}} & 
    \multicolumn{2}{c}{\taskref{elevator}} &
    \multicolumn{1}{c}{Throughput} \\
    \cmidrule(lr){2-3} \cmidrule(lr){4-5} \cmidrule(lr){6-7} 
      & \SR{} & \CR{} & \SR{} & \CR{} & \SR{} & \CR{} & (Hz) \\ \midrule
    \inet{}       & 0 & 53.8 & 30 & 82 & 0 & 50 & 15  \\
    \mfinet{}     & 0 & 72.3 & 20 & 72 & 0 & 50 & 15  \\ 
    \cnnlstm{}    & 0 & 90.1 & 20 & 88 & 0 & 83 & 56 \\ \cmidrule(lr){1-8} 
    \Backbone{}   & 0 & 60.7 & 0 & 58 & 0 & 38 & 125  \\
    \EarlyLayer{} & 0 & 84.6 & 0 & 58 & 20 & 60 & 53 \\ 
    \LateLayer{}  & 0 & 84.8 & 10 & 72 & 70 & 88 & 76 \\
    \textbf{\proposed{}}   & \textbf{90} & \textbf{98.8} & \textbf{90} & \textbf{98} & \textbf{90} &  \textbf{95} & 43 \\
    \human{}      & 90 & 98.9 & 90 & 98 & 100 & 100 & -\\ 
 \bottomrule
\end{tabular}
\vspace{-17pt}
\end{table}

\textbf{Adversarial pedestrian avoidance.} Under our adversarial setup, reactive policies, \ie, \inet{} and \Backbone{}, fail nearly half of the time. This is because they act purely based on the instantaneous position of the pedestrian. \mfinet{} can estimate motion with a short sliding window, but the estimate is too uncertain to achieve consistent success. In comparison, LSTM-like structures have a longer history. The key difference between \proposed{} and \cnnlstm{} is that \proposed{} explicitly reasons both low-level motions, \ie, pedestrian movement, and high-level semantics, \ie, identification of pedestrians, and thus has a higher chance of avoiding the adversary. Missing information at some levels decreases performance, as indicated by the ablation results. 

\textbf{Blind-spot object avoidance.} In this task, reactive agents, \ie, \inet{} and \Backbone{}, have no information on the locations of blind-spot obstacles, but we find that they may succeed by purely mimicking the expert trajectory: Upon seeing the basket, the agent changes its direction to one side of the path and slowly turns back after the obstacle enters the blind spot. The trajectory of turning back may have a curvature small enough to bypass the basket. \mfinet{} has more collisions because it corrects its deviation from the path center faster, which is desirable in obstacle-free environments but increases the chance of failure in this task. With high-level memory, \LateLayer{} outperforms \Backbone{}; with low-level memory, \proposed{} is significantly better than \LateLayer{}. This shows that incorporating memory at a single level may not confer significant advantages, but a memory at multiple spatial scales does. On the other hand, \cnnlstm{} can be viewed as a sequence model where history is only extracted from the highest spatial level, \ie, 1D compressed features, which outperforms \Backbone{}. However, it misses low-level spatial cues, thus suboptimal compared to \proposed{}.

\textbf{Elevator riding.} In this task, agents without memory or with only short memory often exit the elevator at step c), \ie, \inet{} and \mfinet{}. A shallow CNN, \ie, \Backbone{} cannot even accurately recognize the elevator door, and thus has a \CR{} lower than 50\%. \cnnlstm{} extracts temporal information on top-level CNN features, which leads to significantly more success at step c), but we find that it may never exit the elevator door at step d). This indicates that the model may have learned to capture some confounding visual cues, whereas \proposed{} can handle the task well. This shows that the memory needs to be incorporated at the right spatial levels to capture relevant features. Within the \proposed{} model class, adding memory improves performance (\Backbone{} vs. \proposed{}), with \proposed{} achieving the best performance among all policies. This again shows that the multi-scale temporal features confer the most advantages. 

\subsection{Navigation with Multimodal Behaviors}
\label{exp:multimodal}
This subsection focuses on the question: Is our \branch{} structure critical to learning a multimodal policy?

\subsubsection{\textbf{Task}}
We study the case where an agent performs loop closure in a building. The agent is required to make $n = 4$ turns and the objective is to go back to the starting position. The challenge is to learn a policy that shows distinctive behaviors when given different modes.

\subsubsection{\textbf{Baselines}}
To verify our \branch{} design, it is sufficient to compare our model with a variant whose such a structure is ablated.

\subsubsection{\textbf{Results}}
As shown in \tabref{performance:multimodal}, the \branch{} structure can significantly improve the capability of making turns. Note that a high \SR{} is important to perform navigation in practice, because missing turns is costly and could lead to non-recoverable failures.

\begin{table}[t]
\centering
\scriptsize
\caption{navigation with multimodal policy ($N=10$)}
  \label{performance:multimodal}
  \vspace{-7pt}
 \begin{tabular}{l c c c}
    \toprule
    Model & 
    \multicolumn{1}{c}{\SR{}(\%)} & 
    \multicolumn{1}{c}{\CR{}(\%)} &
    \multicolumn{1}{c}{Throughput (Hz)} \\
    \midrule
    \nobranch{}   & 0  & 30 & 43 \\ 
    \textbf{\proposed{}}   & \textbf{90} & \textbf{97.5} & 43 \\
    \human{}      & 100 & 100 & - \\
 \bottomrule
\end{tabular}
\vspace{-15pt}
\end{table}

\begin{figure}
  \centering
  \setlength{\tabcolsep}{6pt} 
    \renewcommand{\arraystretch}{1} 
  \begin{tabular}{cc}
    \includegraphics[width=0.3\columnwidth]{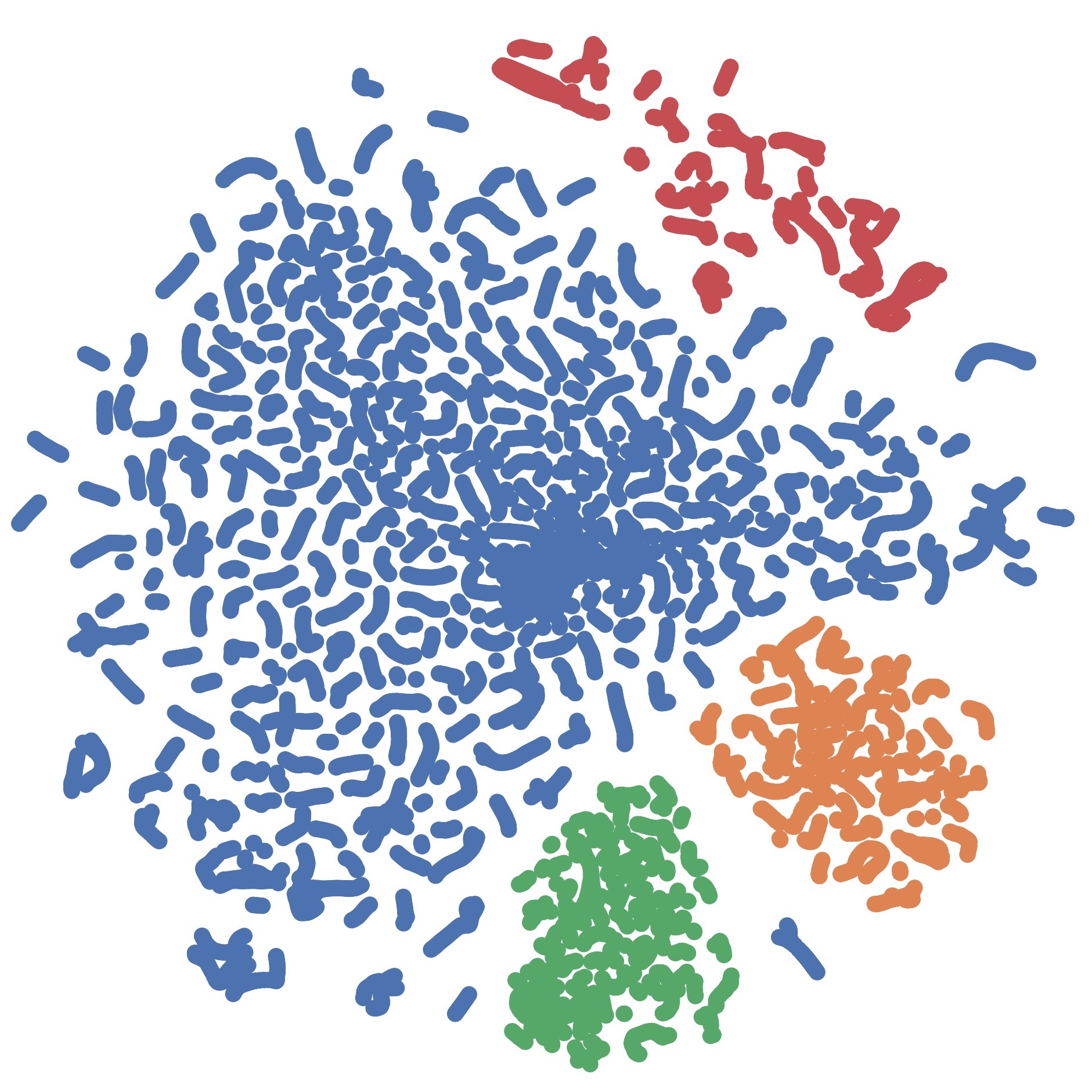} &
    \includegraphics[width=0.3\columnwidth]{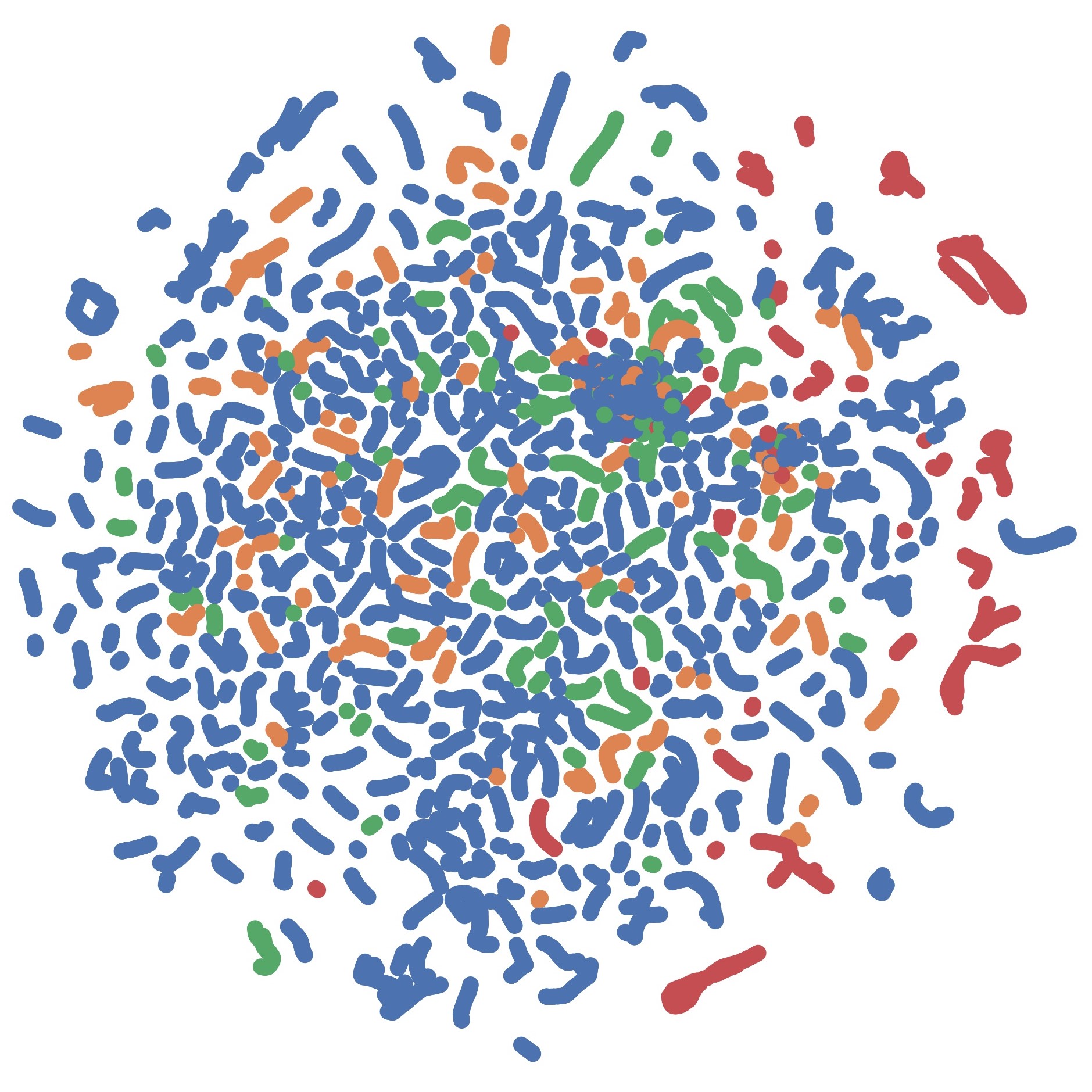} \cr
  \end{tabular}
  \vspace{-8pt}
  \caption{t-SNE visualization of the learned features of \proposed{} (left) and \nobranch{} (right). Blue, green, orange, and red corresponds to the modes \intentforward{}, \intentleft{}, \intentright{}, and \intentelevator{}.}
  \label{tsne}
\vspace{-5pt}
\end{figure}

\begin{figure}[h]
  \centering
  \setlength{\tabcolsep}{6pt} 
    \renewcommand{\arraystretch}{1} 
  \begin{tabular}{cc}
    \includegraphics[width=0.30\columnwidth]{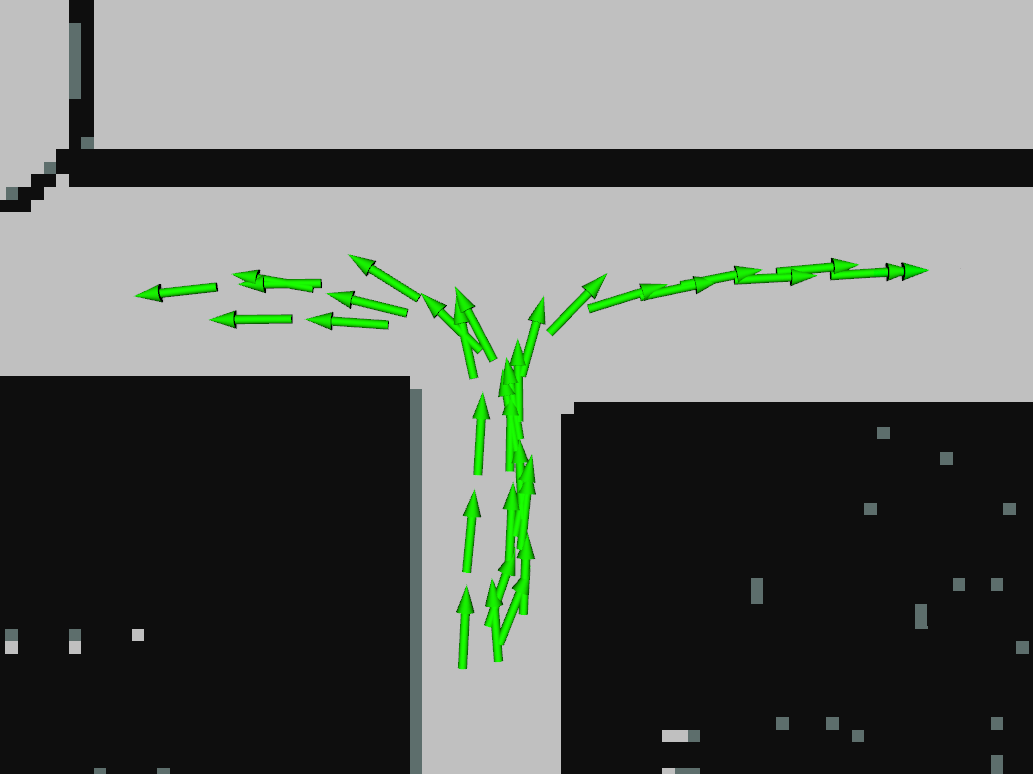} &
    \includegraphics[width=0.30\columnwidth]{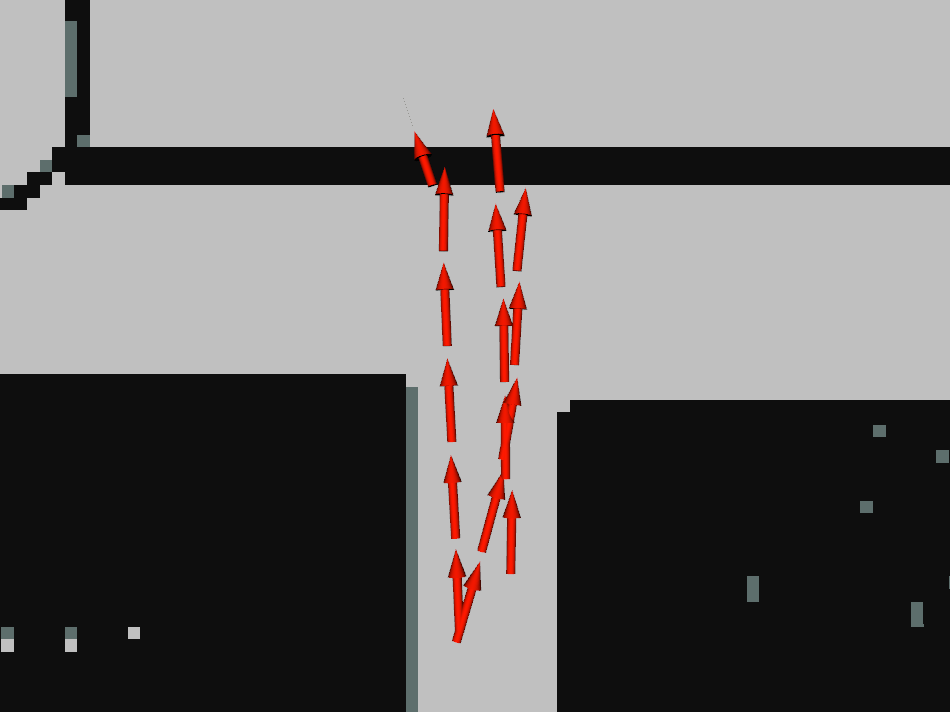} \cr
  \end{tabular}
  \vspace{-2pt}
  \caption{Sample trajectories of \proposed{} (left) and \nobranch{} (right) at a junction. Green and red correspond to success and failure turns. Trajectories are obtained by visual odometry.}
  \label{intersection}
\vspace{-18pt}
\end{figure}

To understand why our structure helps, we project the pooled features before the final MLP (depicted in \figref{archi}) from $\mathbb{R}^{1024}$ to $\mathbb{R}^{2}$ with t-SNE \cite{tsne} (\figref{tsne}). We find that our proposed model learns representations highly separable by modes. As the agent's behaviors are also multimodal, the mapping from the representation space to control space can be more easily learned. In comparison, the baseline features are cluttered and collapse to a unimodal distribution. As illustrated in \figref{intersection}, the baseline model always moves straight forward regardless of mode signals.

In addition, note that the same finding holds true for \intentelevator{}. Here we use turning as a case to reflect the general mode collapse problem under space constraints.

\section{CONCLUSION}
We have presented \proposed{}, a neural motion controller for robot navigation. It is robust to visual variations and partial observations, while maintaining multiple modes of behaviors that are essential to long-horizon navigation. We implemented it on our Spot robot and conducted all experiments in the physical world for relevance to real-world tasks. Importantly, \proposed{} can be adapted to different robot architectures to enable robust local maneuvering, which may simplify robot system design. Our neural network architecture designs may also transfer to other tasks that involve spatio-temporal reasoning or predicting multimodal targets.

Throughout our work, many interesting questions have presented themselves: 
\begin{itemize}
\item 
\textit{How to learn more efficiently?} We have shown that structure has a strong influence on the learning outcome, and we need to discover more
useful structures to scale up robot learning \cite{innate_structure}.
\item \textit{Where to learn?} Learning in the real world is constrained by the amount of data available. To remove the bottleneck, privileged learning and reinforcement learning in the simulation are potential directions, but the simulation-to-reality gap becomes a new challenge.
\item \textit{What can be learned?} It is unclear whether \proposed{} has sufficient representation power or information to learn complex, interactive behaviors that require long-horizon reasoning, in tasks such as crowd driving \cite{lets_drive}.

\end{itemize}






\section*{ACKNOWLEDGMENTS}
This research is supported by the National Research Foundation, Singapore, under its Medium Sized Centre Program, Center for Advanced Robotics Technology Innovation (CARTIN) and by the National University of Singapore (AcRF grant R-252-000-A87-114). B. Ai is supported by an NUS Science and Technology Scholarship.



\clearpage
\bibliography{references}
\bibliographystyle{plain}  
\end{document}